# Artificial intelligence for detection and quantification of rust and leaf miner in coffee crop

ALVARO LEANDRO CAVALCANTE CARNEIRO[1]; LUCAS DE BRITO SILVA[2]; MARISA SILVEIRA ALMEIDA RENAUD FAULIN[3];
[1]São Paulo State Faculty of Technology, Pompéia, São Paulo, Brazil, alvaro.leandro@unesp.br
[2]São Paulo State Faculty of Technology, Pompéia, São Paulo, Brazil, lucas.brito-silva@unesp.br
[3]São Paulo State Faculty of Technology, Pompéia, São Paulo, Brazil, marisa.faulin@fatec.sp.gov.br.

**ABSTRACT**

Pest and disease control plays a key role in agriculture since the damage caused by these agents are responsible for a huge economic loss every year. Based on this assumption, we create an algorithm capable of detecting rust (*Hemileia vastatrix*) and leaf miner (*Leucoptera coffeella*) in coffee leaves (*Coffea arabica*) and quantify disease severity using a mobile application as a high-level interface for the model inferences. We used different convolutional neural network architectures to create the object detector, besides the OpenCV library, k-means, and three treatments: the RGB and value to quantification, and the AFSoft software, in addition to the analysis of variance, where we compare the three methods. The results show an average precision of 81,5% in the detection and that there was no significant statistical difference between treatments to quantify the severity of coffee leaves, proposing a computationally less costly method. The application, together with the trained model, can detect the pest and disease over different image conditions and infection stages and also estimate the disease infection stage.

## 1. INTRODUCTION

Among the various phytosanitary problems found in coffee culture, we can mention bacteria, fungi, viruses, nematodes, and pests. Of these, it is noteworthy the action of rust that has been prominent in coffee since 1869 in East Africa. It generates severe damages, with the early defoliation of coffee trees and production declines in subsequent years. There are reports of a 34% drop in production [1].

Similar to this, the leaf miner began in mid-1850 in Africa, and is currently the most prevalent pest in the world, causing damage to the leaf epidermis, making it malnourished and dry and resulting in losses of up to 40% in production [1].

One way of trying to overcome these phytopathologies is through artificial intelligence, which has had several advances in the last few years, mainly applied to image recognition, not only because of more powerful hardware or data available but also due to the creation of new and better algorithms to solve these problems [2], like convolutional neural networks (CNN) once this network can learn and generalize features from images.

In this work, we used a deep learning-based technique called object detection. This technique is getting more popular [3] and is adopted when the intention is not only to determine whether the object finds or not in the image as most common classification problems but also locate these objects of interest, which is the most visual approach for situations where multiple objects may appear simultaneously in the image, as happens with pests and diseases in leaves, shown in figure 1.

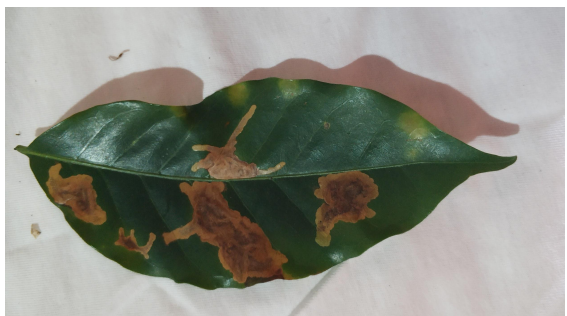

Figure 1. A coffee leaf with miner (brown) and rust (yellow) symptoms

Another essential factor for the farmer is the quantification of disease severity, once through this information, it is possible to control the pathogen earlier and reduce losses. Currently, this work is performed with the aid of diagrammatic scales and descriptive keys and depends on human visual acuity, which is expensive and turns this process very time-consuming.

As an improvement, the use of image analysis methods with computer vision and machine learning is highlighted, among them the k-means, which is an unsupervised learning approach that we choose to perform data clustering based on its similarities and quantify the rust disease.

The AFSoft, a Empresa Brasileira de Pesquisa Agropecuária (EMBRAPA) software, is very common for this analysis in images, mapping, and quantifying diseases that attack plant leaves using Multilayer Perceptron (MLP), a supervised neural network that tries to simulate the human brain [4], and was used as a benchmark with our proposed solution.

The following section presents in more detail the used algorithms, the process through the dataset creation and labeling, techniques used to train and evaluate the model, the preprocessing steps needed for the quantification algorithm, and technologies used to develop the mobile software. Section 3 presents the obtained results while showing improvements that could be done in future research. Finally, in section 4 we conclude the main contributions reached in this paper.

## 2. MATERIALS AND METHODS

### 2.1 Convolutional neural network architectures

The Single-Shot Detector (SSD) is a method for object detection proposed by Liu et al. [5] and used in this work, which highlights its computational efficiency with low memory usage and processing time [6], being able to infer objects in real-time, while maintaining a higher accuracy than the methods proposed previously [5].

An essential part of the SSD training process is the base network, which is a CNN architecture that makes up the first layers of the SSD. The VGG [7] is the one initially used in the detector, but we alternatively chose ResNet and Inception.

ResNet is a residual learning framework proposed by He et al. [8], and its main advantage is the possibility of using deeper neural networks to obtain better results in accuracy, through the creation of residual maps, thus avoiding gradient degradation.

On the other hand, Inception was presented in its first version in 2015 and its second version in 2016 by Szegedy et al. [9], proposing the creation of a new CNN architecture that obtained better results in competitions like the Imagenet Large Scale Visual Recognition Challenge [10] while reducing the number of training parameters, increasing computational viability.

### 2.2 Dataset creation

We randomly collected the leaves with rust and leaf miner at a farmer located in Brazil and the photos were captured in a laboratory in a white background using a smartphone camera in the RGB color system, which is the standard color adopted by digital cameras [11] and presents a satisfactory performance in machine learning concerning other color spaces [12].

The created dataset [13] has 285 and 257 samples of rust and leaf miner, respectively. The images were initially in a resolution of 4000 x 2250 pixels, however, we resized them to 300 x 300 pixels and 512 x 512 pixels due to the SSD standards show by Liu et al. [5] and also by the fact that higher resolutions impact in the training time and computational resources required.

**2.3 Data labeling**

One of the challenges of the object detection technique is the creation of a labeled dataset, where it is necessary to manually specify in the photos the location of the objects of interest through bounding boxes. This process is inefficient, expensive, and time-consuming [3], especially if it is necessary to label dozens of objects per image, require specialized knowledge, or the database is large.

The collected photos, at first, were manually labeled using LabelImage software. However, to facilitate the process mentioned above and speed up the future expansion of the dataset, an algorithm was created to label the images automatically, using a semi-supervised approach. Thus, for each new unlabeled image, the pre-trained model with just a few strongly-labeled pictures infer the objects of interest.

The automatically created labels went through a review process since inherent inaccuracy in detecting objects in some images, which if kept without any post-treatment can reduce the model accuracy, as noted by Gao et al. [14]. Still, the new process adopted is faster and cheaper, as it requires less human effort compared to the standard approach.

**2.4 Data augmentation and transfer learning**

We used data augmentation and transfer learning techniques to improve model performance. Data augmentation creates new images during the training by randomly performing vertical and horizontal inversion, cropping, rescaling, alteration in the position of the bounding boxes, adjustment in contrast, and adjustment in brightness for each image.

The advantages are that it increases the generalizability of the model [15] as it deals with images under different conditions, which has proven significant improvements in results [16], mainly when the number of samples in the database is smaller [5], besides decreasing the chances of overfitting [17, 18].

Transfer learning, on the other hand, improves the results [19] by getting the weights of an already trained model for a couple of days or even weeks in another dataset, usually employing millions of images and different classes. Since this technique improves learning in similar tasks [20], a significant and variate number of classes in the dataset will serve as a general-purpose source for transfer learning. Based on this, we selected a model trained in Microsoft Common Objects in Context (MSCOCO) dataset, proposed by Lin et al. [21], with more than 90 categories and 320,000 images.

**2.5 Training**

The dataset was randomly divided into 80% of the images for training and 20% for testing, as it is a typical data split ratio into machine learning problems [22] and also shows good results concerning other proportions [19].

The evaluation of the object detection model was done by measuring the mean Average Precision (mAP) of all classes known by the model. In Average Precision (AP) as shown by Everingham et al. [23], for a predicted bounding box to be considered true positive, it must have at least a 50% overlap rate, which is represented by the union over intersection (UoI) with the correct bounding box that has been created manually. We can describe this process by Eq.1.

$$UoI = \frac{area\ (Bp \cap Bt)}{area\ (Bp \cup Bt)} \qquad (1)$$

Where *Bp* is the predicted bounding box by the model and *Bt* is the ground truth bounding box.

The construction and parameterization of SSD and the base networks were done in the Tensorflow machine learning system [24]. We trained the model by 2500 epochs in each test, and used the Google Collaboratory platform [25] as the dedicated hardware for the task, through a virtual environment equipped with a Graphic Processing Unit (GPU).

**2.6 Pre-treatment for severity quantification**

As a prerequisite for severity quantification in the platform, the Python language was used with the OpenCV library.

First, to perform the quantification, the original image resolution was decreased by 3.125 times from 4000 × 2250 pixels to 1280 × 720 pixels. This reduction does not affect the final results and also reduces the processing time of the algorithm.

Then, we employed segmentation techniques using the GrabCut method with five iterations, which was also proposed by Rother et al. [26], where only the region of interest (ROI) in the foreground was captured and the remaining pixels were discarded.

Finally, we used the images with two distinct color systems, RGB and HSV, allowing greater exploration of the feature space of our images and obtaining diversified values in terms of severity. Only the value layer of the HSV system was used, which refers to the total brightness of the colors and represents the achromatic part of an image in RGB [27]. Its formula can be represented by the value equation, demonstrated in Eq. 2.

$$Value = max(R', G', B') \qquad (2)$$

At the Eq. 2 $R'$, $B'$ and $G'$ represent the value of the red, blue, and green channels divided by 255 (which is the maximum number of intensities).

**2.7 Severity quantification**

In relation to the severity quantification of the leaf disease, after applying the pretreatment filters, we employed the K-means method with the Sklearn library and tested 3, 4, and 5 clusters. Its pseudocode was represented by Dunham [28].

The algorithm generates a random starting point as a centroid for a given cluster and then tries to assign the other data in the dataset to the clusters according to their proximity using Euclidean distance [29]. It is worth mentioning that the centroid points are updated according to the classifications that the algorithm made, and they stop updating whether the algorithm fulfills its objective, there are few iterations of the data with different clusters or through a fixed number of iterations [28].

The percentage of disease severity was calculated from the following formula (Eq. 3), disregarding the background of the image.

$$DS = \frac{d \times 100}{lad} \qquad (3)$$

whereby, $DS$ is the disease severity, $d$ is disease and $lad$ is the leaf area added to the disease.

**2.8 Treatments of disease quantification**

We used the following treatments: 1) RGB quantification algorithm, 2) Value quantification algorithm and 3) AFSoft software. Visual estimates of rust severity on coffee leaves were used for comparison. In order to obtain the visual estimates, five evaluators trained in disease assessment analyzed 20 diseased coffee leaves and estimated the rust severity using the diagrammatic scale proposed by Cunha et al. [30].

As for proving the severity percentage using the AFSoft tool, having previously separated the images that were used for the tests, some steps were taken: 1) manually selecting what is background, leaf, and disease in three different classes. 2) image binarization (Figure 2). 3) creation of the neural network, if one has not been previously loaded. 4) processing of the selected image or multiple images.

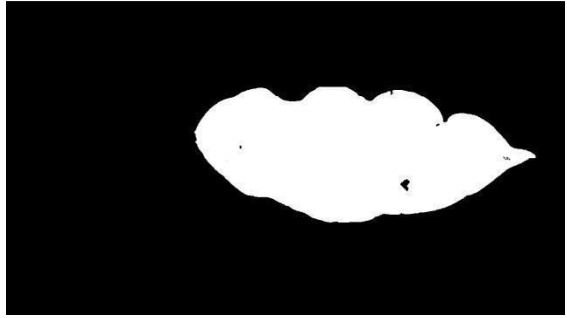

Figure 2. Binarized coffee leaf image using AFSoft

After the results were obtained, statistical tests were performed to prove and substantiate the work. The tests performed were: analysis of variance (ANOVA), confidence intervals, and proportion test, and after the analysis of variance, the three treatments were compared among themselves.

**2.9 Mobile platform development**

We developed a mobile application with the following technologies:

- Ionic: Cross-platform framework used to create the Android and IOS app.
- Node: Provides the development environment for the Application Programming Interface (API).
- Express: Used to the development of a Representational State Transfer (REST) API, generating resources that are consumed by the Ionic models.

The main features of the application were allowing users to capture and store photos without an internet connection, which is a common scenario when collecting leaves samples on a farm.

These photos are later sent to the API which is responsible for dealing with the HTTP requests, storing the pictures to retrain the model with new data, making the inference of the images with the trained model or K-means algorithm, and sending the pictures in the email registered by the user (in case of several photos at once) or returning the processed photo to the application (in case of only one photo). Figure 3 shows how the platform (app + API) works.

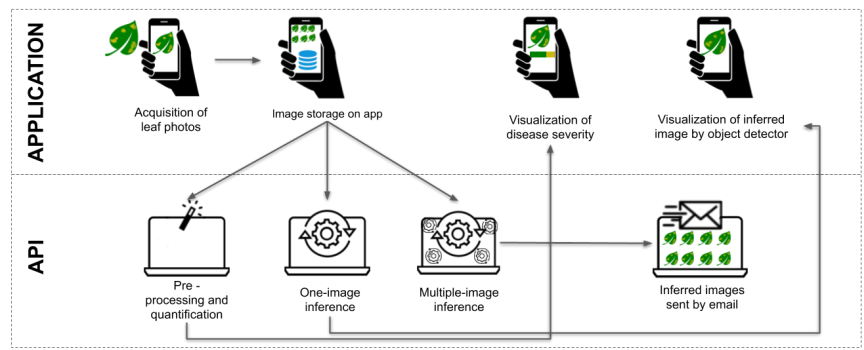

Figure 3. Platform flowchart

## 3. RESULTS AND DISCUSSION

**3.1 Disease and pest detection**

After training and evaluating all base network architectures with the same parameters we can observe that Inception had superior results using SSD with images in a resolution of 512x512 pixels when compared to ResNet, reaching 81.5% of mAP as seen in table 1.

Table 1. mAP obtained by the model in the test set training with Inception and ResNet base networks under different image resolutions.

| Image resolution | ResNet | Inception |
|---|---|---|
| 300 x 300 | 78.3% | 78.9 % |
| 512 x 512 | 67% | 81.5% |

In figure 4 we can see some outcomes of the CNN inference, performing the classification and location of most symptoms present in the respective leaves.

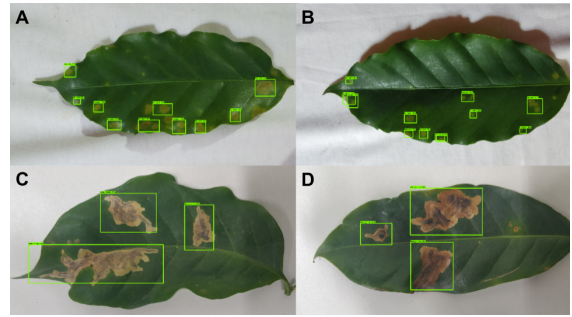

Figure 4. Detection of rust (A, B) and leaf miner (C, D) by the trained model

Similar results were observed by Fuentes et al. [16] in the detection of pests and diseases in tomato crops using not only the SSD but also different architectures and base networks, achieving a maximum mAP of 85,9%.

One way to improve our model at first would be to increase the number of samples since deep learning problems benefit from a larger dataset size, as observed by Sladojevic et al. [18], where training was conducted with more than thirty thousand plant images.

Furthermore, we noticed a higher difficulty for the algorithm to locate the rust disease, even though it has a more significant number of samples. This probably happened because the signs left on the leaf by the rust are considerably smaller than those left by the miner, once SSD has worse performance on small objects [5] and due to the fact these yellowish signs are similar to the eventual light reflections present in the image.

**3.2 Disease quantification**

Among the computational tests, the use of the GrabCut technique stands out, which was very important in the image segmentation stage, since it was possible that only the coffee leaf was captured, thus separating the sick coffee leaf from the background, as can be seen in figure 5.

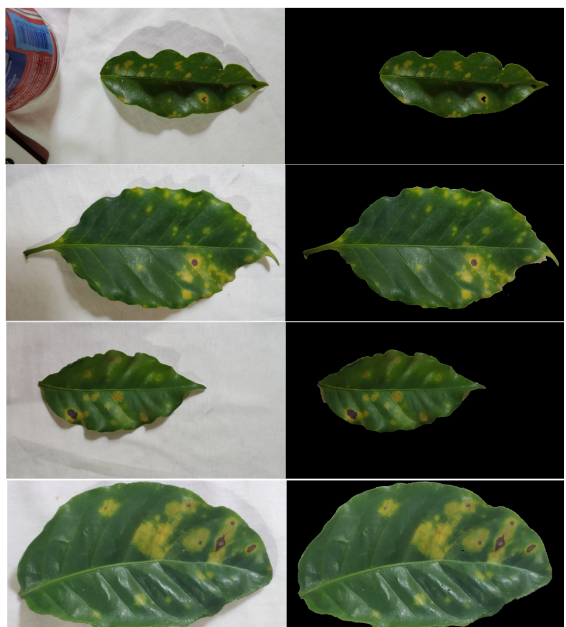

Figure 5. Samples of the GrabCut method result

Another test was performed with AFSoft that automatically discarded the image background, when in black, thus expressing only the value that represents the leaf and the disease. The tool returns, as a result, a table that contains the percentages of leaf and illness as well as a painted image with different colors between disease and leaf (Figure 6).

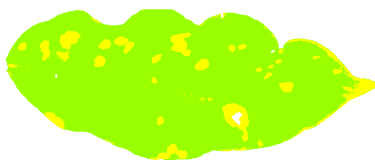

Figure 6. Result image obtained by the AFSoft tool

Therefore, the best results were obtained when using the combination of two techniques, being: I) GrabCut with five iterations, since with this number of iterations it was possible to get precise segmentation results [31] and II) K-Means with five clusters, in the HSV color system with the value channel.

To statistically compare treatments, normality tests were performed, in which the Kolmogorov-Smirnov test was highlighted in the AFSoft results, determining the data as normal for obtaining 0.0169 p-value. Then Tukey's test was conducted from an analysis of variance (table 2) with the factors RGB, value layer, and AFSoft. Thus, with p-value equivalent to 0.399, it was concluded that there was no significant difference between the methods for evaluating the severity of coffee rust, considering a significance level of 5%.

Table 2. Results of the ANOVA test.

|  | DF | Sum Sq. | f-value | f-value | p-value |
|---|---|---|---|---|---|
| Factor | 2 | 39.68028 | 19.84 |  |  |
| Residuals | 57 | 1.210,65 | 21.24 | 0.934 | 0.3989 |

We can observe that from the estimation through a confidence interval (figure 7), the AFSoft obtained discrepant values, however, still without significant difference, as presented in the ANOVA test (Table 2). The upper and lower limits of AFSoft in the confidence interval graph were 9.027 and one of 4.8999, respectively.

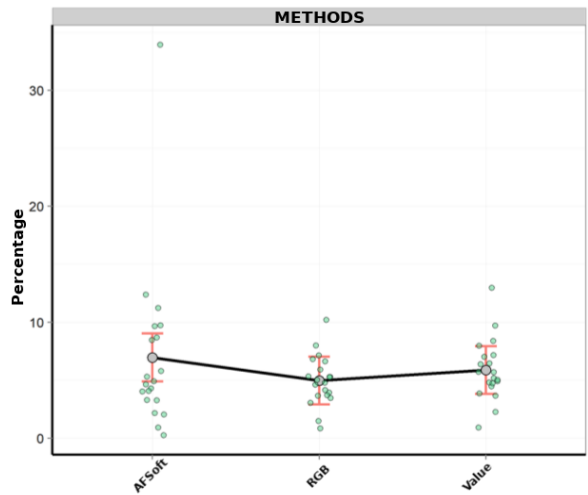

Figure 7. The confidence interval of the means of the AFSoft, RGB, and value treatments

Subsequently, proportion tests were performed between the analyzed methods, thus finding that all processes are similar, since there was no significant difference, from the p-value generated by the normal approximation. To reach this, the methods were submitted to the visual acuity table, thus being able to obtain the proportions of success and failure. Then, with a proportion test of two samples (among all methods), a p-value of 0.735, 0.735, and 1 and a z-index of 0.338, 0.338, and 0 were obtained for the combination of the methods RGB and value layer, RGB and AFSoft and finally, value layer and AFSoft, respectively.

Future advances include the use of a more careful segmentation, color channels that allow greater assertiveness, dynamic scales with less variation, and dynamic clusters for the k-means algorithm.

First, concerning segmentation, tests with other methods are necessary, since the GrabCut technique still has a deficiency to cut figures with low contrast, overlapping of similar colors, or poor adequacy of the fixed rectangle [26], as shown in figure 8.

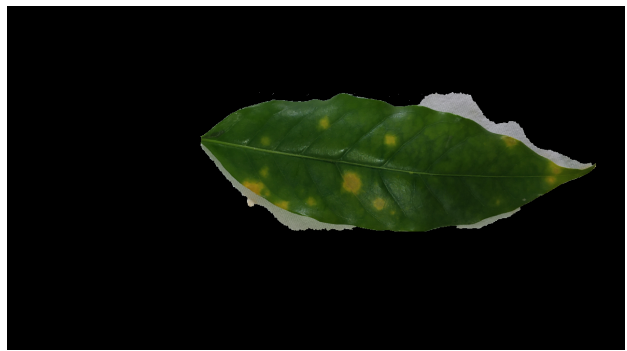

Figure 8. Failure to cut the image using the GrabCut method

Then, the use of a different diagrammatic scale to classify the samples, since the one used by evaluators had a range equivalent to double the percentages, for example, from 12 to 25% and 25% to 50%. This large variety allowed the results to be less thorough.

Finally, we emphasize that from tests carried out in this research, greater assertiveness was observed in diseases with severity between 25 and 50% with the definition of 3 clusters, since when the condition was the third prominent color, its grouping was more significant. However, with fewer symptoms of the disease, 5 clusters obtained a more favorable result, since the signs of the illnesses were smaller and needed a more significant number of clusters to be detected.

## 4. CONCLUSIONS

The deep learning-based object detector proved to be competent in the task of generalizing the characteristics of the images, even with the presence of noise, successfully detecting 81.5% of all symptoms left by these pathogens on the leaves. Furthermore, the proposed rust quantification method achieved similar performance to the AFSoft, but with the advantage of computationally less costly, requiring no prior training of MLP.